\documentclass[runningheads]{llncs}
\usepackage{graphicx}
\usepackage{tikz}
\usepackage{comment}
\usepackage{amsmath,amssymb} % define this before the line numbering.
\usepackage{color}
\usepackage[colorlinks,linkcolor=blue,anchorcolor=blue]{hyperref}

\usepackage[accsupp]{axessibility}  % Improves PDF readability for those with disabilities.

\usepackage[width=122mm,left=12mm,paperwidth=146mm,height=193mm,top=12mm,paperheight=217mm]{geometry}

\usepackage{bm}
\usepackage{algorithm}
\usepackage{algorithmic}
\newcommand{\etal}{\textit{et al}.}
\usepackage{multirow}
\usepackage{booktabs}
\usepackage[normalem]{ulem}
\useunder{\uline}{\ul}{}
\begin{document}
% \linenumbers
\pagestyle{headings}
\mainmatter

\title{RA-Depth: Resolution Adaptive Self-Supervised Monocular Depth Estimation} % Replace with your title

\makeatletter
\newcommand{\printfnsymbol}[1]{%
	\textsuperscript{\@fnsymbol{#1}}%
}
\makeatother

%\begin{comment}
\titlerunning{Resolution Adaptive Monocular Depth Estimation}
\author{Mu He \and Le Hui \and Yikai Bian \and Jian Ren \and Jin Xie\thanks{Corresponding authors.} \and Jian Yang\printfnsymbol{1}}
\authorrunning{M.He, L.Hui, Y.Bian, J.Ren, J.Xie, J.Yang}
\institute{Key Lab of Intelligent Perception and Systems for High-Dimensional Information of Ministry of Education \\ Jiangsu Key Lab of Image and Video Understanding for Social Security \\ PCA Lab, School of Computer Science and Engineering \\ Nanjing University of Science and Technology, China \\
	\email{\{muhe,le.hui,yikai.bian,renjian,csjxie,csjyang\}@njust.edu.cn}}
%\end{comment}
%******************
\maketitle

\begin{abstract}
Existing self-supervised monocular depth estimation methods can get rid of expensive annotations and achieve promising results. However, these methods suffer from severe performance degradation when directly adopting a model trained on a fixed resolution to evaluate at other different resolutions. In this paper, we propose a resolution adaptive self-supervised monocular depth estimation method (RA-Depth) by learning the scale invariance of the scene depth. Specifically, we propose a simple yet efficient data augmentation method to generate images with arbitrary scales for the same scene. Then, we develop a dual high-resolution network that uses the multi-path encoder and decoder with dense interactions to aggregate multi-scale features for accurate depth inference. Finally, to explicitly learn the scale invariance of the scene depth, we formulate a cross-scale depth consistency loss on depth predictions with different scales. Extensive experiments on the KITTI, Make3D and NYU-V2 datasets demonstrate that RA-Depth not only achieves state-of-the-art performance, but also exhibits a good ability of resolution adaptation. Source code is available at \url{https://github.com/hmhemu/RA-Depth}.
\end{abstract}

\section{Introduction}
\label{sec:introduction}
Monocular depth estimation~\cite{make3d,eigen2014depth,laina2016deeper,liu2015learning,ladicky2014pulling,fu2018deep} recovers a pixel-wise depth map from a single image, which is a challenging but essential task in computer vision. It has promoted the development of various applications such as autonomous driving, robot navigation, 3D scene reconstruction, and augmented reality. Due to the difference of acquisition devices in different application scenarios, the resolutions of the acquired images are different. Thus, to reduce resource consumption, how to make a single trained model of monocular depth estimation adaptive to the changes of image resolution in a self-supervised manner is a valuable and open problem.

\begin{figure}[t]
	\centering
	\setlength{\abovecaptionskip}{0.12cm}
	\includegraphics[width=1.0\linewidth]{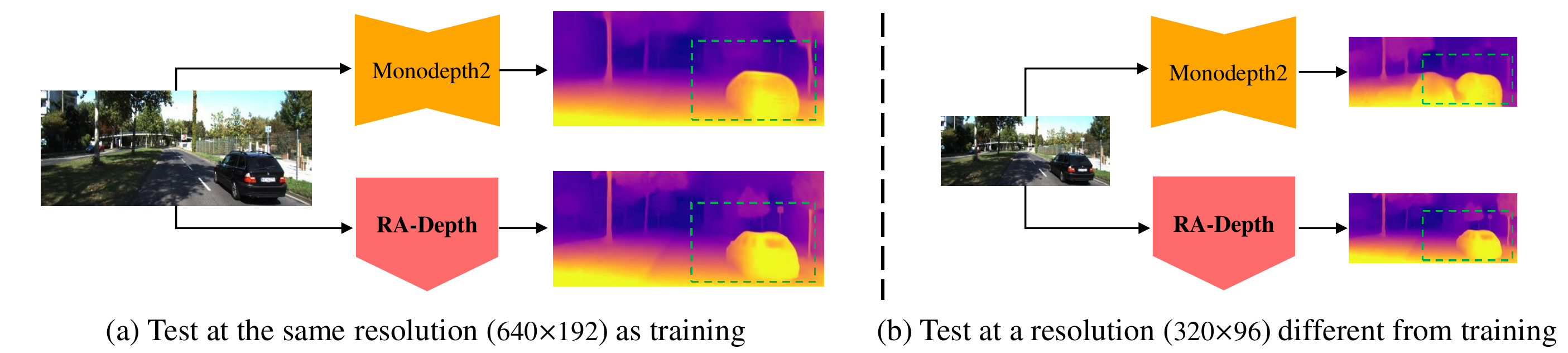}
	
	\caption{Depth predictions for different resolutions. We compare the depth predictions of Monodepth2~\cite{monodepth2} and our RA-Depth at different resolutions. Both Monodepth2 and RA-Depth are trained at the fixed resolution of $640\times192$, and then tested at $640\times192$ (a) and $320\times96$ (b). RA-Depth exhibits the good resolution adaptation ability when tested at a resolution different from training.} \label{fig_intro}
\end{figure}

Existing self-supervised monocular depth estimation methods~\cite{monodepth,3dunsupervised,zhou2017,monodepth2,HRDepth} usually use geometrical constraints on stereo pairs or monocular sequences as the sole source of supervision and have made great progress.
However, these supervision methods cannot adapt to the changes of image resolution at inference, whose performance drops severely when the test resolution is inconsistent with the training resolution.
As shown in Fig.~\ref{fig_intro}, we compare the depth prediction results of existing advanced work Monodepth2~\cite{monodepth2} and our RA-Depth.
Both Monodepth2 and RA-Depth are trained at the fixed resolution of $640\times192$ and then tested at two resolutions $640\times192$ and $320\times96$.
It can be seen that Monodepth2 predicts inconsistent and poor depth maps when tested at different resolutions. To test at different resolutions, existing methods train an individual model for each resolution. As a result, this not only requires unavoidable training overhead and high storage cost, but also limits the flexibility and practicality of monocular depth estimation models.

In this paper, we propose a resolution adaptive self-supervised monocular depth estimation method named as RA-Depth.
First, we propose an arbitrary-scale data augmentation method to generate images with arbitrary scales for the same scene.
Specifically, we change the scale of the input image while maintaining the image resolution/size (\textit{i.e.}, width$\times$height) through randomly resizing, cropping, and stitching (details in Section~\ref{sec:aug}).
These generated images with arbitrary scales can mimic the scale variations at different image resolutions, thus prompting the model to implicitly learn the scale invariance of scene depth.
Then, we propose an efficient monocular depth estimation framework with multi-scale feature fusion. In our framework, we use the superior high-resolution representation network HRNet~\cite{hrnet} as encoder, and design a decoder with efficient multi-scale feature fusion.
As a result, the encoder and decoder form a dual HRNet architecture, which can fully exploit and aggregate multi-scale features to infer the accurate depth.
Thanks to this efficient design, we can achieve better depth estimation performance with the lower network overhead.
Finally, we propose a novel cross-scale depth consistency loss to explicitly learn the scale invariance of scene depth. Thus, the monocular depth estimation model can predict the consistent depth maps for the same scene, even if the scales of input images are different. Extensive experiments on the KITTI~\cite{kitti}, Make3D~\cite{make3d}, and NYU-V2~\cite{nyuv2} datasets demonstrate that RA-Depth achieves state-of-the-art performance on monocular depth estimation and shows a good ability of resolution adaptation.

To summarize, the main contributions of our work are as follows:
\begin{itemize}
\item To the best of our knowledge, our RA-Depth is the first work to tackle the image resolution adaptation for self-supervised monocular depth estimation.

\item We propose an arbitrary-scale data augmentation method to promote the model to learn depths from images with different scales.

\item We develop an efficient dual high-resolution network with multi-scale feature fusion for monocular depth estimation, which is trained with a novel cross-scale depth consistency loss.

\item Extensive experiments demonstrate that our RA-Depth achieves state-of-the-art performance on monocular depth estimation, and has a good generalization ability to different resolutions.
\end{itemize}

\section{Related Work}
\subsection{Supervised Monocular Depth Estimation}
Monocular depth estimation aims to predict the scene depth from single color image. Early works~\cite{eigen2014depth,liu2015deep,laina2016deeper,xu2018structured,hu2019revisiting,lee2019big,yin2019enforcing} estimated continuous depth maps pixel by pixel using convolutional neural networks by minimizing the error between the prediction and ground-truth depths. Furthermore, the geometry~\cite{yin2019enforcing,patil2022p3depth} or multi-task constraints~\cite{qi2018geonet,zhang2019pattern} are imposed on the depth estimation network to predict more accurate depth maps. However, these continuous regression methods usually suffer from slow convergence and unsatisfactory local solutions. Deep ordinal regression network~\cite{fu2018deep} thus introduced a spacing-increasing discretization strategy to discretize depth, where the depth estimation problem is converted into the ordinal depth regression problem. AdaBins~\cite{bhat2021adabins} proposed a Transformer based depth estimation scheme by dividing the depth range into different bins. Recently, some works~\cite{parida2021beyond,leistner2022towards} employed the multimodel learning for depth estimation and further improved its performance.

\subsection{Self-Supervised Monocular Depth Estimation}
Self-supervised methods convert the monocular depth estimation task into an image reconstruction problem, and they use the photometric loss on stereo pairs or monocular sequences as the supervisory signal to train networks.

One type of self-supervised monocular depth estimation methods employ the geometric constraints on stereo pairs to learn depth. Garg \etal~\cite{garg2016unsupervised} first propose a self-supervised monocular depth estimation framework, where depth is estimated from synchronized stereo image pairs.
Following this framework, Monodepth~\cite{monodepth} designs a left-right disparity consistency loss to produce more accurate prediction results, while 3Net~\cite{poggi2018learning} introduces a trinocular stereo assumption to solve the occlusion problem between stereo pairs.
Based on these methods, Depth Hints~\cite{depthhints} uses the off-the-shelf Semi-Global Matching (SGM) algorithm~\cite{sgm2005,sgm2007} to obtain complementary depth hints, which can enhance existing photometric loss and guide the network to learn more accurate weights. Recently, EPCDepth~\cite{epcdepth} adopted a data grafting technique to make the model focus on non-vertical image positions for accurate depth inference.

Another type of depth estimation methods with self-supervision come from unlabelled monocular videos.
SfM-Learner~\cite{zhou2017} is a pioneering work that jointly learns monocular depth and ego-motion from monocular videos in a self-supervised way.
Based on SfM-Learner, SC-SfMLearner~\cite{bian2019} proposes a geometry consistency loss to ensure the scale-consistency of depth. Monodepth2~\cite{monodepth2} introduces a minimum reprojection loss and an auto-masking loss to solve the problem of occlusions and moving objects. Poggi \etal~\cite{Poggi_CVPR_2020} explore the uncertainty of self-supervised monocular depth estimation and propose a novel peculiar technique specifically designed for self-supervised depth estimation. Besides, many recent works~\cite{chen2019towards,sgddepth,safenet,zhu2020edge,guizilini2020semantically} utilize extra semantic supervision to aid self-supervised monocular depth estimation.
More recently, some works~\cite{HRDepth,diffnet} have developed high-performance monocular depth estimation models through efficient network architecture designs.

\section{Method}
\label{sec:method}
\subsection{Problem Formulation}
Self-supervised methods exploit geometric constraints on stereo pairs or monocular videos as the supervision signal for training.
%Our proposed framework uses unlabelled monocular video for training, where each target frame $\bm{I_t}$ corresponds to two source source frames $\bm{I_s}$ ($\bm{s}\in \{\bm{t-1}, \bm{t+1}\}$) that adjacent to $\bm{I_t}$ temporally.
In this paper, our proposed framework uses unlabelled monocular videos for training, each training instance contains a target frame $\bm{I}_t$ and two source frames $\bm{I}_s$ ($s\in \{t-1, t+1\}$) adjacent to $\bm{I}_t$ temporally.
With the depth $\bm{D}_t$ of the target view and the camera pose $\bm{T}_{t\rightarrow s}$ between the target view and source view, the image synthesis process from the source view to the target view can be expressed as 
$\bm{I}_{s\rightarrow t} = \bm{I}_s\langle proj(\bm{D}_t, \bm{T}_{t\rightarrow s}, \bm{K})\rangle$, where $\bm{K}$ represents the known camera intrinsics, $\langle \rangle$ is the sampling operator~\cite{stn} and $proj()$ is the coordinate projection operation~\cite{zhou2017}.
The photometric error is composed of $L_1$ and SSIM~\cite{ssim}:
\begin{equation}
\label{eq2}
pe(\bm{I}_a, \bm{I}_b) = \frac{\alpha}{2}(1 - SSIM(\bm{I}_a, \bm{I}_b)) + (1 - \alpha)\left \| \bm{I}_a - \bm{I}_b \right \| _1.
\end{equation}
Following~\cite{monodepth2}, in order to handle the issue of occlusion, we adopt the per-pixel minimum reprojection loss as our photometric loss:
\begin{equation}
\label{eq3}
\mathcal{L}_{ph}(\bm{I}_t, \bm{I}_{s\rightarrow t}) = \mathop{min}\limits_{s} pe(\bm{I}_t, \bm{I}_{s\rightarrow t}).
\end{equation}
The edge-aware smoothness loss is used to deal with disparity discontinuities:
\begin{equation}
\label{eq4}
\mathcal{L}_{sm}(\bm{D}_t, \bm{I}_t) = \left| \partial_x \bm{D}_t^*\right| e^{-\left|  \partial_x \bm{I}_t \right|}  + \left| \partial_y \bm{D}_t^*\right| e^{-\left|  \partial_y \bm{I}_t \right|}
\end{equation}
where $\partial_x$ and $\partial_y$ are the gradients in the horizontal and vertical direction respectively. Besides, $\bm{D}_t^* = \bm{D}_t / \overline {\bm{D}_t}$ is the mean-normalized inverse depth from~\cite{monodepth2} to discourage shrinking of the estimated depth.

\begin{figure}[t]
	\centering
	\includegraphics[scale=0.42]{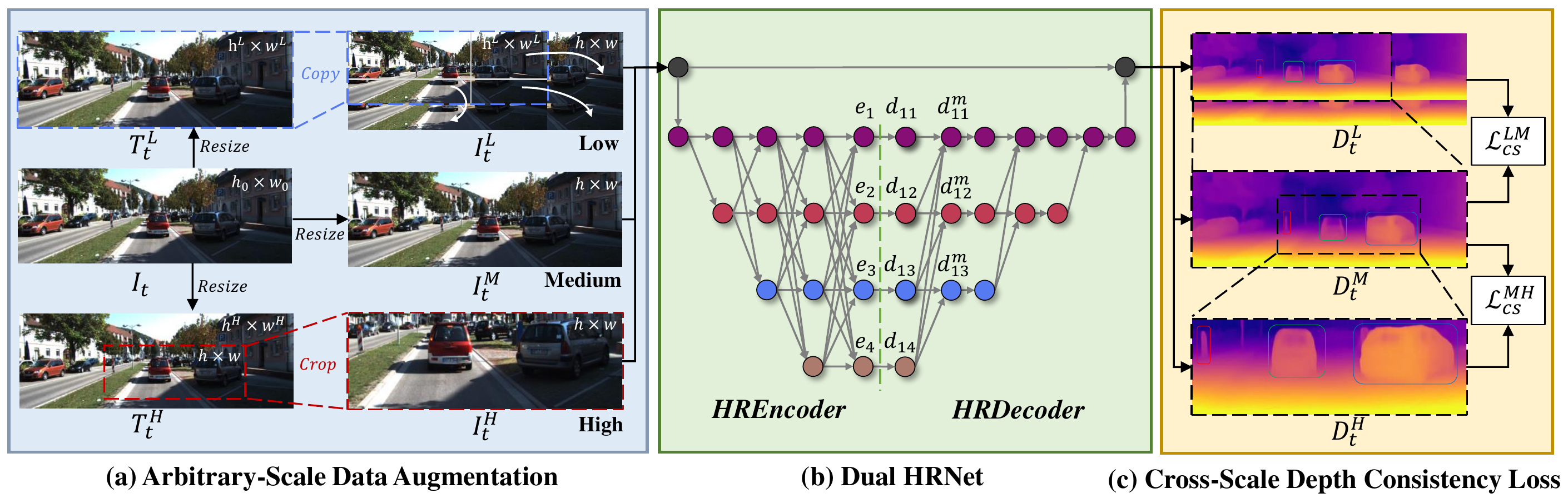}
	\caption{Overview of RA-Depth. (a) First, given an original instance \{$\bm{I}_t$, $\bm{I}_s$\} ($s\in \{t-1, t+1\}$, $t$ for target image and $s$ for source image), we use our arbitrary-scale data augmentation to generate three training instances with different scales: \{$\bm{I}_t^L$, $\bm{I}_s^L$\}, \{$\bm{I}_t^M$, $\bm{I}_s^M$\}, and \{$\bm{I}_t^H$, $\bm{I}_s^H$\}. For simplicity, only target images $\bm{I}_t^L$, $\bm{I}_t^M$, and $\bm{I}_t^H$ are shown in this figure. (b) Then, we use our proposed efficient Dual HRNet to separately predict depth maps ($\bm{D}_t^L$, $\bm{D}_t^M$, $\bm{D}_t^H$) for these target images. Note that Dual HRNet only inputs a single target image and then outputs the corresponding depth map at a time, and all target images share the same Dual HRNet. (c) Finally, we compute the proposed cross-scale depth consistency losses $\mathcal{L}_{cs}^{LM}$ (between $\bm{D}_t^L$ and $\bm{D}_t^M$) and $\mathcal{L}_{cs}^{MH}$ (between $\bm{D}_t^M$ and $\bm{D}_t^H$) to constrain the network to learn the scale invariance of the scene depth.} \label{fig_overview}
\end{figure}

\subsection{Resolution Adaptive Self-Supervised Framework}
\subsubsection{Overview.}
\label{sec:overview}
An overview of our proposed method can be seen in Fig.~\ref{fig_overview}.
Given an original instance \{$\bm{I}_t$, $\bm{I}_s$\} ($s\in \{t-1, t+1\}$, $t$ for target image and $s$ for source image), we employ our proposed arbitrary-scale data augmentation to expand three training instances: \{$\bm{I}_t^L$, $\bm{I}_s^L$\}, \{$\bm{I}_t^M$, $\bm{I}_s^M$\}, and \{$\bm{I}_t^H$, $\bm{I}_s^H$\}, where \textit{L}, \textit{M} and \textit{H} denote the low, middle and high scales, respectively.
For simplicity, we only show $\bm{I}_t^L$, $\bm{I}_t^M$, and $\bm{I}_t^H$ in Fig.~\ref{fig_overview}(a). Then, we use our proposed efficient Dual HRNet to separately predict depth maps ($\bm{D}_t^L$, $\bm{D}_t^M$, $\bm{D}_t^H$) of these target views, and use the pose network to predict the relative 6D camera pose $\bm{T}_{t\rightarrow s}$ from the target view to the source view.
With $\bm{D}_t^L$ and $\bm{T}_{t\rightarrow s}$, we can synthesize image ${{\tilde{\bm{I}}}_{s\rightarrow t}^L}$ by interpolating the source image $\bm{I}_s^L$.
In the same way, we can synthesize images ${{\tilde{\bm{I}}}_{s\rightarrow t}^M}$ and ${{\tilde{\bm{I}}}_{s\rightarrow t}^H}$.
Then we compute the photometric losses between real images ($\bm{I}_t^L$, $\bm{I}_t^M$, $\bm{I}_t^H$) and synthesized images (${{\tilde{\bm{I}}}_{s\rightarrow t}^L}$, ${{\tilde{\bm{I}}}_{s\rightarrow t}^M}$, ${{\tilde{\bm{I}}}_{s\rightarrow t}^H}$).
Note that all image synthesis processes use the same camera pose $\bm{T}_{t\rightarrow s}$ predicted from $\bm{I}_t^M$ to $\bm{I}_s^M$.
Finally, the proposed cross-scale depth consistency losses $\mathcal{L}_{cs}^{LM}$ (between $\bm{D}_t^L$ and $\bm{D}_t^M$) and $\mathcal{L}_{cs}^{MH}$ (between $\bm{D}_t^M$ and $\bm{D}_t^H$) are computed. For clarity, the pose network, photometric loss, and smoothness loss are not shown in Fig.~\ref{fig_overview}.

\subsubsection{Arbitrary-Scale Data Augmentation.}
\label{sec:aug}
The details of arbitrary-scale data augmentation for a single input are shown in Algorithm~\ref{alg1}.
For a single original image $\bm{I}$ of shape (\textit{c}, \textit{$h_0$}, \textit{$w_0$}), we first obtain three images ($\bm{T}^L$, $\bm{T}^M$, and $\bm{T}^H$) with different resolutions by the $Resize$ operation. The resolution of $\bm{T}^M$ is fixed to (\textit{h}, \textit{w}), which denotes the middle scale. The resolutions of $\bm{T}^L$ and $\bm{T}^H$ vary randomly over a continuous range, controlled by the scale factors $s^L$ and $s^H$, respectively. The ranges of $s^L$ and $s^H$ are [0.7, 0.9] and [1.1, 2.0], respectively, making the resolution of $\bm{T}^L$ lower than (\textit{h}, \textit{w}) and the resolution of $\bm{T}^H$ higher than (\textit{h}, \textit{w}).
Then we can generate a low scale image $\bm{I}^L$ from $\bm{T}^L$ by means of image stitching (represented by white arrows in Fig.~\ref{fig_overview}(a)), and directly treat $\bm{T}^M$ as the middle scale image $\bm{I}^M$, and generate the high scale image $\bm{I}^H$ from $\bm{T}^H$ by means of $RandomCrop$ (represented by red dashed box in Fig.~\ref{fig_overview}(a)).
Finally, we obtain three images $\bm{I}^L$, $\bm{I}^M$, and $\bm{I}^H$ with the same shape of (\textit{c}, \textit{h}, \textit{w}) at different scales, which served as the input of the network. Note, in this paper, (\textit{c}, \textit{$h_0$}, \textit{$w_0$}) is (3, 375, 1242), and (\textit{c}, \textit{h}, \textit{w}) is (3, 192, 640).

\begin{algorithm}[t]
	\caption{ Arbitrary-Scale Data Augmentation.} 
	\label{alg1} 
	\begin{algorithmic}[1]
		\REQUIRE 
		An original image $\bm{I}$ of shape (\textit{c}, \textit{$h_0$}, \textit{$w_0$}).
		\ENSURE
		Three training images $\bm{I}^L$, $\bm{I}^M$, $\bm{I}^H$ of shape (\textit{c}, \textit{h}, \textit{w}).
		\STATE Random initializing the scale factors $s^L$ and $s^H$ with respective ranges of [0.7, 0.9] and [1.1, 2.0]; Zero initializing $\bm{I}^L$, $\bm{I}^M$, $\bm{I}^H$ of shape (\textit{c}, \textit{h}, \textit{w});
		\label{ code:fram:initializing }
		
		\STATE $(h^L, w^L), (h^M, w^M), (h^H, w^H) = int((h, w) \times s^L), (h, w), int((h, w) \times s^H)$;
		
		\STATE Generating the low-resolution image ($\bm{I}^L$);\\
		(a) Resizing the original image $\bm{I}$ \\
		\quad\quad $\bm{T}^L \gets Resize(\bm{I})$ to $(c, h^L, w^L)$ \\
		
		(b) Slicing and copying $\bm{T}^L$\\
		\quad\quad $\bm{I}^L[:, :h^L, :w^L] \gets \bm{T}^L[:, :, :]$ \\ 
		
		%\quad (c) Stitching to the bottom region \\
		\quad\quad $\bm{I}^L[:, h^L:h, :w^L] \gets \bm{T}^L[:, (2 \times h^L - h) :h^L, :w^L]$ \\
		
		%\quad (d) Stitching to the right region \\
		\quad\quad $\bm{I}^L[:, :h^L, w^L:w] \gets \bm{T}^L[:, :h^L, (2 \times w^L - w) :w^L]$ \\ 
		
		%\quad (e) Stitching to the bottom right region \\
		\quad\quad $\bm{I}^L[:, h^L:h, w^L:w] \gets \bm{T}^L[:, (2 \times h^L - h) :h^L, (2 \times w^L - w) :w^L]$ \\
		
		\STATE Generating the middle-resolution image ($\bm{I}^M$); \\
		(a) Resizing the original image $\bm{I}$ \\
		\quad\quad $\bm{T}^M \gets Resize(\bm{I})$ to $(c, h^M, w^M)$\\
			
		(b) Copying $\bm{T}^M$\\
		\quad\quad $\bm{I}^M[:, :, :] \gets \bm{T}^M[:, :, :]$
		
		\STATE Generating the high-resolution image ($\bm{I}^H$);\\
		(a) Resizing the original image $\bm{I}$ \\
		\quad\quad $\bm{T}^H \gets Resize(\bm{I})$ to $(c, h^H, w^H)$\\
		
		(b) Randomly cropping $\bm{T}^H$\\
		\quad\quad $\bm{I}^H[:, :, :] \gets RandomCrop (\bm{T}^H)$ to $(c, h, w)$
 
		%\RETURN
		\STATE Return $\bm{I}^L$, $\bm{I}^M$, $\bm{I}^H$;
	\end{algorithmic}
\end{algorithm}

\subsubsection{Dual HRNet for Monocular Depth Estimation.}
\label{sec:network}
As shown in Fig.~\ref{fig_overview}(b), we develop an efficient dual high-resolution network (Dual HRNet) with multi-scale feature fusion for monocular depth estimation.
We use the superior high-resolution representation network HRNet18~\cite{hrnet} as the encoder, named HREncoder. The core idea of HRNet~\cite{hrnet} is to perform efficient multi-scale feature fusion while maintaining high-resolution feature representations, so as to obtain semantically richer and spatially more precise multi-scale features.
Inspired by HRNet~\cite{hrnet}, we design a decoder with efficient multi-scale feature fusion named HRDecoder.
Specifically, HRDecoder first inherits the multi-scale features of the encoder, and then gradually fuses low-scale features while maintaining the high-resolution feature representations.
Let $\bm{e}_i$ denote the feature of the encoder at the $i$-th level, $\bm{d}_{ji}$ denote the feature of the decoder at the $i$-th level and the $j$-th stage, and $\bm{d}^m_{ji}$ denote the feature after multi-scale feature fusion at the $i$-th level and the $j$-th stage.
Let us take the calculation of $\bm{d}_{1i}$ and $\bm{d}^m_{1i}$ (refer to Fig.~\ref{fig_overview}(b)) at the first stage as an example:
\begin{equation}
\label{eq5}
\begin{cases}
\bm{d}_{1i} = \operatorname{CONV}_{3\times3}(e_i), & \textit{i} = 1,2,3,4 \\  
\bm{d}^m_{1i} = \bm{d}_{1i} + [\operatorname{CONV}_{1\times1}(\mu(\bm{d}_{1k}))] & \textit{i} = 1,2,3,\textit{k} = i+1, \ldots, 4
\end{cases}
\end{equation}
where $\operatorname{CONV}_{3\times3}$ represents a $3\times3$ convolution layer, and $\operatorname{CONV}_{1\times1}$ represents an $1\times1$ convolution layer. In addition, $\mu(\cdot)$ is an upsampling operator, and $[\cdot]$ is a summation operator. By exploiting the HRDecoder, the multi-scale features can be fully fused from the low-resolution feature map to the high-resolution feature map. It is desired that the Dual HRNet can effectively learn depth information of different scales for accurate depth estimation at different test resolutions.
%By exploiting the HRDecoder, the multi-scale features can be fully fused from the low-resolution feature maps to the high-resolution feature maps.
%can fully exploit and aggregate multi-scale features to infer the accurate depth.

\subsubsection{Cross-Scale Depth Consistency Loss.}
\label{sec:loss}
As shown in Fig.~\ref{fig_overview}(c), we compute the cross-scale depth consistency losses $\mathcal{L}_{cs}^{ML}$ (between $\bm{D}_t^M$ and $\bm{D}_t^L$) and $\mathcal{L}_{cs}^{MH}$ (between $\bm{D}_t^M$ and $\bm{D}_t^H$) to constrain the model to explicitly learn the scale invariance of scene depth. For example, the same pillar or car among $\bm{D}_t^L$, $\bm{D}_t^M$ and $\bm{D}_t^H$ should have the same depth as shown in Fig.~\ref{fig_overview}(c).
Specifically, according to the known scale factors $s^L$ and $s^H$ as well as the cropped position of \textit{RandomCrop} in Algorithm~\ref{alg1}, we can obtain the pixel correspondence among $\bm{D}_t^L$, $\bm{D}_t^M$ and $\bm{D}_t^H$.
For $\bm{D}_t^M$ and $\bm{D}_t^H$, we first find the corresponding position of $\bm{D}_t^H$ in $\bm{D}_t^M$, which is represented by the black dashed box (named as ${\tilde{\bm{D}}_t^M}$) in $\bm{D}_t^M$.
Then we resize $\bm{D}_t^H$ to keep the same size as ${\tilde{\bm{D}}_t^M}$ to obtain ${\tilde{\bm{D}}_t^H}$. Finally, we calculate the scene depth consistency loss between ${\tilde{\bm{D}}_t^M}$ and ${\tilde{\bm{D}}_t^H}$:
\begin{equation}
\label{eq6}
\mathcal{L}_{cs}^{MH} = \frac{\alpha}{2}(1 - SSIM({\tilde{\bm{D}}_t^M}, {\tilde{\bm{D}}_t^H})) + (1 - \alpha)\left \| {\tilde{\bm{D}}_t^M} - {\tilde{\bm{D}}_t^H} \right \| _1.
\end{equation}
Similarly, for $\bm{D}_t^L$ and $\bm{D}_t^M$, we first find the corresponding position of $\bm{D}_t^M$ in $\bm{D}_t^L$, which is represented by the black dashed box (named as ${\hat{\bm{D}}_t^L}$) in $\bm{D}_t^L$.
Then we resize $\bm{D}_t^M$ to keep the same size as ${\hat{\bm{D}}_t^L}$ to obtain ${\hat{\bm{D}}_t^M}$.
Finally, we calculate the cross-scale depth consistency loss between ${\hat{\bm{D}}_t^L}$ and ${\hat{\bm{D}}_t^M}$:
\begin{equation}
\label{eq7}
\mathcal{L}_{cs}^{LM} = \frac{\alpha}{2}(1 - SSIM({\hat{\bm{D}}_t^L}, {\hat{\bm{D}}_t^M})) + (1 - \alpha)\left \| {\hat{\bm{D}}_t^L} - {\hat{\bm{D}}_t^M} \right \| _1.
\end{equation}

\subsubsection{Final Training Loss.}
Our overall loss function can be formulated as follows:
\begin{equation}
\label{eq8}
\begin{aligned}
\mathcal{L}_{f} = &\gamma(\mathcal{L}_{ph}(\bm{I}_t^L, {\tilde{\bm{I}}_{s\rightarrow t}^L}) + \mathcal{L}_{ph}(\bm{I}_t^M, {\tilde{\bm{I}}_{s\rightarrow t}^M}) + \mathcal{L}_{ph}(\bm{I}_t^H, {\tilde{\bm{I}}_{s\rightarrow t}^H}))\\
&+ \lambda(\mathcal{L}_{sm}({\tilde{\bm{D}}_t^L}, \bm{I}_t^L) + \mathcal{L}_{sm}({\tilde{\bm{D}}_t^M}, \bm{I}_t^M) + \mathcal{L}_{sm}({\tilde{\bm{D}}_t^H}, \bm{I}_t^H))\\
&+ \beta(\mathcal{L}_{cs}^{LM} + \mathcal{L}_{cs}^{MH})
\end{aligned}
\end{equation}
Here, $\gamma$, $\lambda$, and $\beta$ are the hyper-parameters of the photometric loss, smoothness loss, and our proposed scene depth consistency loss.

\section{Experiments}
\label{sec:experiments}
%In this section, we compare our proposed method RA-Depth to existing state-of-the-art self-supervised monocular depth estimation methods on the KITTI \cite{kitti}, Make3D \cite{make3d}, and NYU-V2 \cite{nyuv2} datasets.

\subsection{Implementation Details}
Our proposed method is implemented in PyTorch \cite{pytorch} and trained on a single Titan RTX GPU.
Our models are trained for 20 epochs using Adam optimizer \cite{kingma2014adam}, with a batch size of 12.
The learning rate for the first 15 epochs is $10^{-4}$ and then dropped to $10^{-5}$  for the last 5 epochs.
The hyper-parameters $\gamma$, $\lambda$, and $\beta$ of the final training loss in Eq.~(\ref{eq8}) are set to 1.0, 0.001, and 1.0, respectively.
We set $\alpha$ = 0.85 in Eq.~(\ref{eq2}),  Eq.~(\ref{eq6}), and Eq.~(\ref{eq7}).
For the monocular depth estimation network, We implement our proposed Dual HRNet described in Section \ref{sec:network}, which uses HRNet18 \cite{hrnet} as the encoder and performs efficient multi-scale feature fusion on the decoder.
For the pose estimation network, we use the same architecture as Monodepth2~\cite{monodepth2}, which contains ResNet18~\cite{resnet} followed by several convolutional layers.
Similar to \cite{monodepth2,diffnet}, we use the weights pretrained on ImageNet \cite{imagenet} to initialize HRNet18 and ResNet18.
During training, the input resolution of networks is 640 $\times$ 192.
To improve the training speed, we only output a single-scale depth for the depth estimation network and compute the loss on the single-scale depth, instead of computing losses on multi-scale outputs of the depth estimation network (4 scales in \cite{zhou2017,monodepth2} and 6 scales in \cite{ranjan2019competitive}).

\begin{table}[t]
	\centering
	\caption{Quantitative monocular depth estimation results on the KITTI dataset~\cite{kitti} using the Eigen split~\cite{eigen2014depth}. We divide compared methods into three categories based on training data. In each category, the best results are in \textbf{bold} and the second are \underline{underlined}. The test resolution is the same as the training resolution for all these methods. S: trained on stereo pairs; M: trained on monocular videos; Se: trained with semantic labels.}\label{tab1}
	\renewcommand\tabcolsep{3.0pt}
	\resizebox{0.97\textwidth}{!}{
		\begin{tabular}{l|c|c|cccc|ccc}
			\toprule[1.5pt]
			\multicolumn{1}{c|}{\multirow{2}{*}{Method}} & \multirow{2}{*}{Train} & \multirow{2}{*}{Resolution} & \multicolumn{4}{c|}{Error Metric $\downarrow$}                                 & \multicolumn{3}{c}{Accuracy Metric $\uparrow$}              \\ \cline{4-10} 
			&                        &                     & AbsRel              & SqRel              & RMSE              & RMSElog              & $\delta\textless{}1.25$              & $\delta\textless{}1.25^2$              & $\delta\textless{}1.25^3$              \\ \hline\hline
			Monodepth~\cite{monodepth}              & S                      & 512$\times$256             & 0.148          & 1.344          & 5.927          & 0.247          & 0.803          & 0.922          & 0.964          \\
			3Net~\cite{poggi2018learning}                     & S                      & 512$\times$256             & 0.142          & 1.207          & 5.702          & 0.240          & 0.809          & 0.928          & 0.967          \\
			Chen~\cite{chen2019towards}                     & S+Se                   & 512$\times$256             & 0.118          & 0.905          & 5.096          & 0.211          & 0.839          & 0.945          & 0.977          \\
			Monodepth2~\cite{monodepth2}               & S                      & 640$\times$192             & 0.109          & 0.873          & 4.960          & 0.209          & 0.864          & 0.948          & 0.975          \\
			Depth Hints~\cite{depthhints}              & S                      & 640$\times$192             & {\ul 0.106}    & {\ul 0.780}    & {\ul 4.695}    & {\ul 0.193}    & {\ul 0.875}    & {\ul 0.958}    & {\ul 0.980}    \\
			EPCDepth~\cite{epcdepth}                 & S                      & 640$\times$192             & \textbf{0.099} & \textbf{0.754} & \textbf{4.490} & \textbf{0.183} & \textbf{0.888} & \textbf{0.963} & \textbf{0.982} \\ \hline
			Zhan~\cite{zhan2018unsupervised}                     & M+S                     & 608$\times$160             & 0.135          & 1.132          & 5.585          & 0.229          & 0.820          & 0.933          & 0.971          \\
			EPC++~\cite{luo2019every}                    & M+S                     & 832$\times$256             & 0.128          & 0.935          & 5.011          & 0.209          & 0.831          & 0.945          & 0.979          \\
			Monodepth2~\cite{monodepth2}               & M+S                     & 640$\times$192             & 0.106          & 0.818          & 4.750          & 0.196          & 0.874          & 0.957          & 0.979          \\
			HR-Depth~\cite{HRDepth}                 & M+S                     & 640$\times$192             & 0.107          & 0.785          & {\ul 4.612}    & {\ul 0.185}    & {\ul 0.887}    & {\ul 0.962}    & {\ul 0.982}    \\
			Depth Hints~\cite{depthhints}              & M+S                     & 640$\times$192             & {\ul 0.105}    & {\ul 0.769}    & 4.627          & 0.189          & 0.875          & 0.959          & {\ul 0.982}    \\
			DIFFNet~\cite{diffnet}                  & M+S                     & 640$\times$192             & \textbf{0.101} & \textbf{0.749} & \textbf{4.445} & \textbf{0.179} & \textbf{0.898} & \textbf{0.965} & \textbf{0.983} \\ \hline
			SfMLearner~\cite{zhou2017}               & M                      & 640$\times$192             & 0.183          & 1.595          & 6.709          & 0.270          & 0.734          & 0.902          & 0.959          \\
			EPC++~\cite{luo2019every}                    & M                      & 832$\times$256             & 0.141          & 1.029          & 5.350          & 0.216          & 0.816          & 0.941          & 0.976          \\
			SC-SfMLearner~\cite{bian2019}            & M                      & 832$\times$256             & 0.114          & 0.813          & 4.706          & 0.191          & 0.873          & 0.960          & 0.982          \\
			Monodepth2~\cite{monodepth2}               & M                      & 640$\times$192             & 0.115          & 0.903          & 4.863          & 0.193          & 0.877          & 0.959          & 0.981          \\
			SGDDepth~\cite{sgddepth}                 & M+Se                   & 640$\times$192             & 0.113          & 0.835          & 4.693          & 0.191          & 0.879          & 0.961          & 0.981          \\
			SAFENet~\cite{safenet}                  & M+Se                   & 640$\times$192             & 0.112          & 0.788          & 4.582          & 0.187          & 0.878          & 0.963          & {\ul 0.983}          \\
			PackNet-SfM~\cite{packnet}              & M                      & 640$\times$192             & 0.111          & 0.785          & 4.601          & 0.189          & 0.878          & 0.960          & 0.982          \\
			Mono-Uncertainty~\cite{Poggi_CVPR_2020}         & M                      & 640$\times$192             & 0.111          & 0.863          & 4.756          & 0.188          & 0.881          & 0.961          & 0.982          \\
			HR-Depth~\cite{HRDepth}                 & M                      & 640$\times$192             & 0.109          & 0.792          & 4.632          & 0.185          & 0.884          & 0.962          & {\ul 0.983}          \\
			DIFFNet~\cite{diffnet}                  & M                      & 640$\times$192             & {\ul 0.102}    & {\ul 0.764}    & {\ul 4.483}    & {\ul 0.180}    & {\ul 0.896}    & {\ul 0.965}    & {\ul 0.983}    \\
			\textbf{RA-Depth}                 & M                      & 640$\times$192             & \textbf{0.096} & \textbf{0.632} & \textbf{4.216} & \textbf{0.171} & \textbf{0.903} & \textbf{0.968} & \textbf{0.985} \\ 
			\bottomrule[1.5pt]
	\end{tabular}}
\end{table}

\subsection{Results}

\subsubsection{Basic Results.}
We evaluate the monocular depth estimation performance of our single model on the KITTI dataset \cite{kitti} using the data split of Eigen \etal~\cite{eigen2014depth}.
Following Zhou \etal~\cite{zhou2017}, we remove static frames, using 39810, 4424, and 697 images for training, validation, and test, respectively.
During the evaluation, we cap depth to 80m and adopt the same conventional metrics as \cite{eigen2014depth}.
Since we use monocular videos for training, we perform median scaling introduced by \cite{zhou2017} on our predicted depths when evaluating.
Table~\ref{tab1} shows the basic quantitative results and all models are tested at the same resolution as training.
Our proposed method outperforms all existing self-supervised monocular depth estimation approaches significantly.
In particular, RA-Depth achieves the SqRel error of 0.632, the error reduction rate of 17\% over the best-published result 0.764.

\begin{table}[h]
	\centering
	\caption{Resolution adaptive comparison results on the KITTI dataset~\cite{kitti} using the Eigen split~\cite{eigen2014depth}.
		All these models are trained at the resolution of 640$\times$192 and then tested at four different resolutions including 416$\times$128, 512$\times$160, 832$\times$256, and 1024$\times$320.}\label{tab2}
	\renewcommand\tabcolsep{3.0pt}
	\resizebox{0.97\textwidth}{!}{
		\begin{tabular}{l|c|cccc|ccc}
			\toprule[1.5pt]
			\multicolumn{1}{c|}{\multirow{2}{*}{Method}} & \multirow{2}{*}{\begin{tabular}[c]{@{}c@{}}Test\\ Resolution \end{tabular}} & \multicolumn{4}{c|}{Error Metric $\downarrow$}                                 & \multicolumn{3}{c}{Accuracy metric $\uparrow$}              \\ \cline{3-9} 
			&                                                                     & AbsRel              & SqRel              & RMSE              & RMSElog              & $\delta\textless{}1.25$              & $\delta\textless{}1.25^2$              & $\delta\textless{}1.25^3$              \\ \hline\hline
			Monodepth2~\cite{monodepth2}              & 416$\times$128                                                             & 0.184          & 1.365          & 6.146          & 0.268          & 0.719          & 0.911          & 0.966          \\
			HR-Depth~\cite{HRDepth}                 & 416$\times$128                                                             & 0.157          & 1.120          & 5.669          & 0.231          & 0.787          & 0.938          & 0.976          \\
			DIFFNet~\cite{diffnet}                  & 416$\times$128                                                             & {\ul 0.142}    & {\ul 1.068}    & {\ul 5.552}    & {\ul 0.218}    & {\ul 0.826}    & {\ul 0.948}    & {\ul 0.979}    \\
			\textbf{RA-Depth}        & 416$\times$128                                                             & \textbf{0.111} & \textbf{0.723} & \textbf{4.768} & \textbf{0.187} & \textbf{0.874} & \textbf{0.961} & \textbf{0.984} \\ \hline
			Monodepth2~\cite{monodepth2}               & 512$\times$160                                                             & 0.132          & 1.062          & 5.318          & 0.209          & 0.846          & 0.951          & 0.978          \\
			HR-Depth~\cite{HRDepth}                 & 512$\times$160                                                             & 0.128          & {\ul 0.978}    & {\ul 4.999}    & 0.199          & 0.855          & 0.957          & {\ul 0.981}    \\
			DIFFNet~\cite{diffnet}                  & 512$\times$160                                                             & {\ul 0.121}    & 1.007          & 5.107          & {\ul 0.196}    & {\ul 0.870}    & {\ul 0.958}    & {\ul 0.981}    \\
			\textbf{RA-Depth}        & 512$\times$160                                                             & \textbf{0.101} & \textbf{0.658} & \textbf{4.373} & \textbf{0.175} & \textbf{0.895} & \textbf{0.967} & \textbf{0.985} \\ \hline
			Monodepth2~\cite{monodepth2}               & 832$\times$256                                                             & 0.131          & 0.992          & 5.451          & 0.222          & 0.815          & 0.947          & 0.978          \\
			HR-Depth~\cite{HRDepth}                 & 832$\times$256                                                             & 0.136          & 0.920          & {\ul 5.321}    & 0.218          & 0.817          & 0.949          & {\ul 0.980}    \\
			DIFFNet~\cite{diffnet}                  & 832$\times$256                                                             & {\ul 0.128}    & {\ul 0.896}    & 5.328          & {\ul 0.214}    & {\ul 0.834}    & {\ul 0.950}    & {\ul 0.980}    \\
			\textbf{RA-Depth}        & 832$\times$256                                                             & \textbf{0.095} & \textbf{0.613} & \textbf{4.106} & \textbf{0.170} & \textbf{0.906} & \textbf{0.969} & \textbf{0.985} \\ \hline
			Monodepth2~\cite{monodepth2}               & 1024$\times$320                                                            & 0.193          & 1.335          & 6.058          & 0.271          & 0.673          & 0.921          & 0.972          \\
			HR-Depth~\cite{HRDepth}                 & 1024$\times$320                                                            & 0.183          & 1.241          & 6.054          & 0.267          & 0.687          & 0.920          & 0.972          \\
			DIFFNet~\cite{diffnet}                  & 1024$\times$320                                                            & {\ul 0.163}    & {\ul 1.153}    & {\ul 6.011}    & {\ul 0.251}    & {\ul 0.743}    & {\ul 0.928}    & {\ul 0.974}    \\
			\textbf{RA-Depth}        & 1024$\times$320                                                            & \textbf{0.097} & \textbf{0.608} & \textbf{4.131} & \textbf{0.174} & \textbf{0.901} & \textbf{0.968} & \textbf{0.985} \\ 
			\bottomrule[1.5pt]
	\end{tabular}}
\end{table}

\subsubsection{Resolution Adaptation Results.}
\label{sec:resluts_radepth}
\begin{figure}[t]
	\centering
	\includegraphics[scale=0.39]{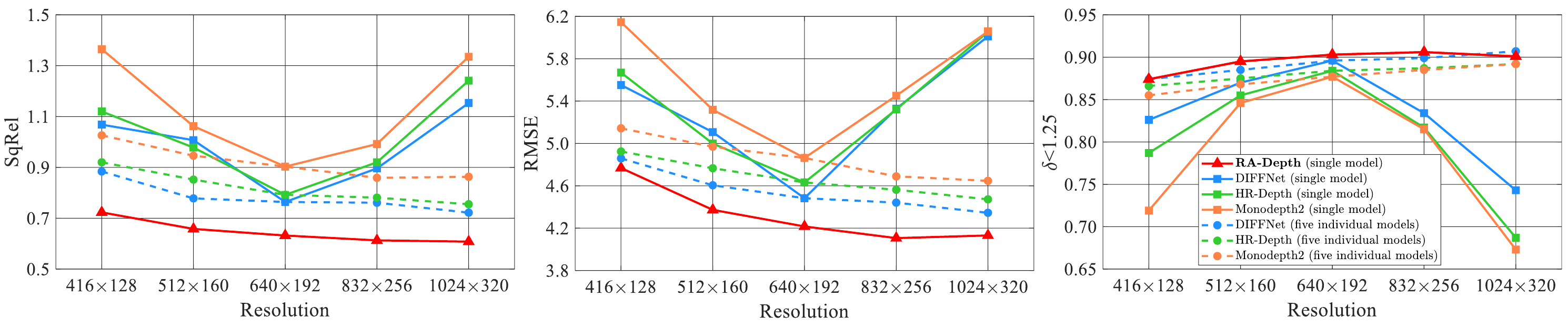}
	\caption{Resolution adaptation results on the error metrics (SqRel, RMSE) and the accuracy metric ($\delta\textless{}1.25$). Solid lines denote the resolution adaptation results tested at five different resolutions using a single model trained on 640$\times$192.
		Dotted lines denote results tested at five different resolutions using five individual models trained on each test resolution.
		Our single model RA-Depth outperforms other single models significantly on all metrics and is comparable to these individual models trained separately at each test resolution. This figure is best viewed in a zoomed-in document.} \label{fig_lines}
\end{figure}

\begin{figure}[h]
	\centering
	\includegraphics[width=0.97\linewidth]{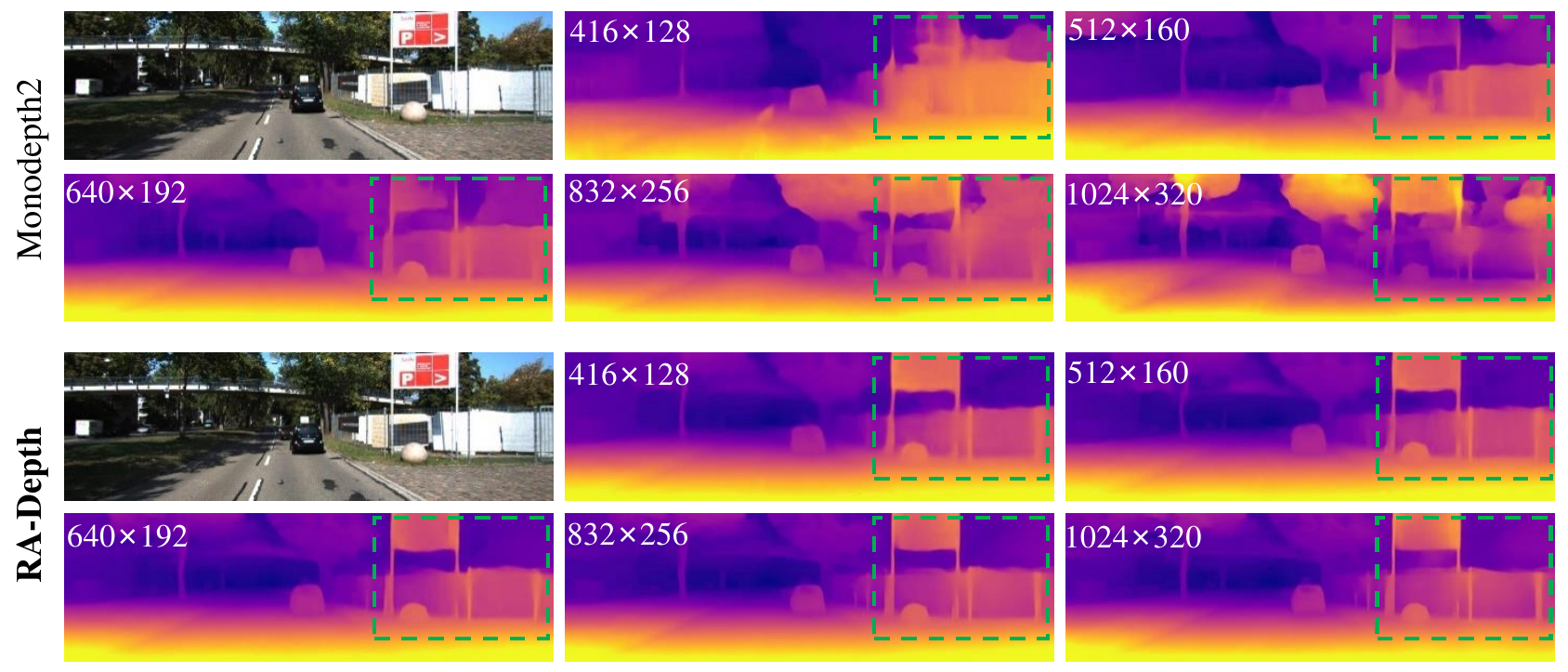}
	\caption{Visualization results of resolution adaptation. We use models trained at the resolution of 640$\times$192 to predict depths at various test resolutions.} \label{fig_depth_visuals}
\end{figure}

As shown in Table~\ref{tab2}, we compare our single model RA-Depth with existing three advanced approaches including Monodepth2 \cite{monodepth2}, HR-Depth \cite{HRDepth}, and DIFFNet \cite{diffnet} in terms of resolution adaptation on the KITTI dataset~\cite{kitti} using the Eigen split~\cite{eigen2014depth}.
All these models are trained at the resolution of 640$\times$192 and then tested at four different resolutions including 416$\times$128, 512$\times$160, 832$\times$256, and 1024$\times$320.
Note, we directly resize the original image to generate four
resolution images without using the scale augmentation strategy in the training process.
The results show that existing methods perform poorly in terms of resolution adaptation.
That is to say, existing methods cannot adapt to the changes of image resolution during testing. In contrast, our single model RA-Depth maintains high performance across different test resolutions, significantly outperforming existing self-supervised methods.
Fig.~\ref{fig_lines} also shows the resolution adaptation results of different models on two error metrics SqRel and RMSE and one accuracy metric $\delta\textless{}1.25$.
It can be seen that our single model RA-Depth achieves the lowest error on two error metrics at all resolutions, even lower than the errors of these individual models trained separately at each test resolution.
Moreover, our single model achieves the highest accuracy at all resolutions compared with other single models and is comparable to these individual models.
Fig.~\ref{fig_depth_visuals} illustrates the visualization results of our RA-Depth and Monodepth2~\cite{monodepth2} for resolution adaptation in the depth estimation. It can be observed that the depth maps predicted by the single model Monodepth2 are inconsistent at different resolutions (see green rectangles). In contrast, our single model RA-Depth shows credible and consistent depth predictions at different resolutions, which demonstrates that RA-Depth can learn the scale-invariance of scene depth.

\begin{table}[h]
	\centering
	\caption{Generalization results on the Make3D~\cite{make3d} and NYU-V2~\cite{nyuv2} datasets. All models are trained at the resolution of $640\times192$ on the KITTI dataset~\cite{kitti}. The upper part of this table reports the generalization results at the same resolution as training. The bottom half of this table reports the average results tested at four different resolutions (416$\times$128, 512$\times$160, 832$\times$256, 1024$\times$320), which represent the results of resolution adaptation across datasets.}\label{tab_make3d_nyuv2}
	\renewcommand\tabcolsep{3.0pt}
	\resizebox{0.97\textwidth}{!}{
		\begin{tabular}{llccccccc}
			\toprule[1.5pt]
			\multicolumn{1}{l|}{\multirow{2}{*}{Dataset}}      & \multicolumn{1}{c|}{\multirow{2}{*}{Method}} & \multicolumn{4}{c|}{Error Metric $\downarrow$}                                                      & \multicolumn{3}{c}{Accuracy Metric $\uparrow$}              \\ \cline{3-9} 
			\multicolumn{1}{c|}{}                              & \multicolumn{1}{l|}{}                         & AbsRel              & SqRel              & RMSE              & \multicolumn{1}{c|}{RMSElog}              & $\delta\textless{}1.25$              & $\delta\textless{}1.25^2$              & $\delta\textless{}1.25^3$             \\ \hline\hline
			\multicolumn{9}{c}{{(a) Results At The Same Resolution As Training}}                                                                                                                                                                                   \\ \hline
			\multicolumn{1}{l|}{\multirow{4}{*}{Make3D}}       & \multicolumn{1}{l|}{Monodepth2~\cite{monodepth2}}              & 0.322          & 3.589          & 7.418          & \multicolumn{1}{c|}{0.163}          & 0.559          & 0.799          & 0.909          \\
			\multicolumn{1}{l|}{}                              & \multicolumn{1}{l|}{HR-Depth~\cite{HRDepth}}                & 0.315          & 3.208          & 7.024          & \multicolumn{1}{c|}{0.159}          & 0.562          & 0.798          & 0.913          \\
			\multicolumn{1}{l|}{}                              & \multicolumn{1}{l|}{DIFFNet~\cite{diffnet}}                 & {\ul 0.309}    & {\ul 3.313}    & {\ul 7.008}    & \multicolumn{1}{c|}{{\ul 0.155}}    & {\ul 0.575}    & {\ul 0.815}    & {\ul 0.919}    \\
			\multicolumn{1}{l|}{}                              & \multicolumn{1}{l|}{\textbf{RA-Depth}}       & \textbf{0.277} & \textbf{2.703} & \textbf{6.548} & \multicolumn{1}{c|}{\textbf{0.143}} & \textbf{0.612} & \textbf{0.846} & \textbf{0.933} \\ \hline
			\multicolumn{1}{l|}{\multirow{4}{*}{NYU-V2}} & \multicolumn{1}{l|}{Monodepth2~\cite{monodepth2}}              & 0.377          & 0.778          & 1.388          & \multicolumn{1}{c|}{0.414}          & 0.442          & 0.728          & 0.879          \\
			\multicolumn{1}{l|}{}                              & \multicolumn{1}{l|}{HR-Depth~\cite{HRDepth}}                & 0.321          & 0.521          & 1.150          & \multicolumn{1}{c|}{0.367}          & 0.490          & 0.774          & 0.913          \\
			\multicolumn{1}{l|}{}                              & \multicolumn{1}{l|}{DIFFNet~\cite{diffnet}}                 & {\ul 0.307}    & {\ul 0.481}    & {\ul 1.095}    & \multicolumn{1}{c|}{{\ul 0.352}}    & {\ul 0.518}    & {\ul 0.796}    & {\ul 0.921}    \\
			\multicolumn{1}{l|}{}                              & \multicolumn{1}{l|}{\textbf{RA-Depth}}       & \textbf{0.250} & \textbf{0.286} & \textbf{0.843} & \multicolumn{1}{c|}{\textbf{0.293}} & \textbf{0.605} & \textbf{0.861} & \textbf{0.955} \\ \hline
			\multicolumn{9}{c}{{(b) Average Results Of Resolution Adaptation At Four Different Resolutions}}                                                                                                                                                                                        \\ \hline
			\multicolumn{1}{l|}{\multirow{4}{*}{Make3D}}       & \multicolumn{1}{l|}{Monodepth2~\cite{monodepth2}}              & 0.342          & 3.862          & 7.921          & \multicolumn{1}{c|}{0.172}          & 0.509          & 0.776          & 0.901          \\
			\multicolumn{1}{l|}{}                              & \multicolumn{1}{l|}{HR-Depth~\cite{HRDepth}}                & 0.340          & {\ul 3.527}    & {\ul 7.577}    & \multicolumn{1}{c|}{0.171}          & 0.502          & 0.773          & 0.903          \\
			\multicolumn{1}{l|}{}                              & \multicolumn{1}{l|}{DIFFNet~\cite{diffnet}}                 & {\ul 0.333}    & 3.753          & 7.700          & \multicolumn{1}{c|}{{\ul 0.167}}    & {\ul 0.528}    & {\ul 0.790}    & {\ul 0.909}    \\
			\multicolumn{1}{l|}{}                              & \multicolumn{1}{l|}{\textbf{RA-Depth}}       & \textbf{0.278} & \textbf{2.722} & \textbf{6.655} & \multicolumn{1}{c|}{\textbf{0.146}} & \textbf{0.601} & \textbf{0.844} & \textbf{0.932} \\ \hline
			\multicolumn{1}{l|}{\multirow{4}{*}{NYU-V2}} & \multicolumn{1}{l|}{Monodepth2~\cite{monodepth2}}              & 0.380          & 0.778          & 1.384          & \multicolumn{1}{c|}{0.413}          & 0.445          & 0.730          & 0.882          \\
			\multicolumn{1}{l|}{}                              & \multicolumn{1}{l|}{HR-Depth~\cite{HRDepth}}                & 0.325          & 0.520          & {\ul 1.147}    & \multicolumn{1}{c|}{0.377}          & 0.491          & 0.778          & {\ul 0.915}    \\
			\multicolumn{1}{l|}{}                              & \multicolumn{1}{l|}{DIFFNet~\cite{diffnet}}                 & {\ul 0.322}    & {\ul 0.508}    & 1.167          & \multicolumn{1}{c|}{{\ul 0.364}}    & {\ul 0.500}    & {\ul 0.781}    & 0.913          \\
			\multicolumn{1}{l|}{}                              & \multicolumn{1}{l|}{\textbf{RA-Depth}}       & \textbf{0.265} & \textbf{0.326} & \textbf{0.890} & \multicolumn{1}{c|}{\textbf{0.304}} & \textbf{0.586} & \textbf{0.848} & \textbf{0.948} \\ 
			\bottomrule[1.5pt]
		\end{tabular}}
\end{table}

%\begin{figure}
%	\centering
%	\includegraphics[scale=0.365]{figs/depths_make3d_nyuv2.pdf}
%	\caption{Visualization results on the Make3D~\cite{make3d} and NYU-V2~\cite{nyuv2} datasets. All models are trained on the KITTI dataset \cite{kitti}.} \label{fig_depth_make3d}
%\end{figure}

\subsubsection{Generalization Results Across Datasets.}

In Table~\ref{tab_make3d_nyuv2}, we report the generalization performance on the Make3D~\cite{make3d} and NYU-V2 ~\cite{nyuv2} datasets using models trained at the resolution of $640\times192$ on the KITTI dataset~\cite{kitti}.
For a fair comparison, we adopt the same evaluation criteria, cropping approach \cite{monodepth2} and data preprocessing strategy \cite{IndoorSfMLearner} for all models.
When the test resolution is the same as the training resolution ($640\times192$), we achieve state-of-the-art results whether generalizing to the outdoor dataset Make3D or the indoor dataset NYU-V2.
Moreover, we also test our model at four different resolutions (416$\times$128, 512$\times$160, 832$\times$256, 1024$\times$320) and take the average results as the resolution adaptation results across datasets.
These results demonstrate that our model still has the good resolution adaptation ability when tested across datasets.
%See more additional qualitative results on supplementary materials.
%Qualitative results on the Make3D and NYU-V2 datasets can be seen in Fig.~\ref{fig_depth_make3d}.

\begin{table}[h]
	\centering
	\caption{Ablation results for each component of our method.
		All models are trained at the resolution of $640\times192$ on the KITTI dataset~\cite{kitti} using the Eigen split~\cite{eigen2014depth}.
		AS-Aug: arbitrary-scale data augmentation. D-HRNet: Dual HRNet. CS-Loss: cross-scale depth consistency loss. pt: pretrained weights on the ImageNet dataset. (a) We test the effects of each component on overall performance at the same resolution as training. (b) We report the average results tested at four different resolutions (416$\times$128, 512$\times$160, 832$\times$256, 1024$\times$320) to verify the effectiveness of each component on the resolution adaptation ability.}
	\label{tab_ablation}
	\renewcommand\tabcolsep{3.0pt}
	\resizebox{0.98\textwidth}{!}{
		\begin{tabular}{lccccccccccc}
			\toprule[1.5pt]
			\multicolumn{1}{l|}{\multirow{2}{*}{}}        & \multicolumn{1}{c|}{\multirow{2}{*}{Pretrained}} & \multicolumn{1}{c|}{\multirow{2}{*}{AS-Aug}} & \multicolumn{1}{c|}{\multirow{2}{*}{D-HRNet}} & \multicolumn{1}{c|}{\multirow{2}{*}{CS-Loss}} & \multicolumn{4}{c|}{Error Metric $\downarrow$}                                                      & \multicolumn{3}{c}{Accuracy Metric $\uparrow$}                   \\ \cline{6-12} 
			\multicolumn{1}{l|}{}                         & \multicolumn{1}{c|}{}                            & \multicolumn{1}{c|}{}                        & \multicolumn{1}{c|}{}                         & \multicolumn{1}{c|}{}                         & AbsRel              & SqRel              & RMSE                                  & \multicolumn{1}{c|}{RMSElog}             & $\delta\textless{}1.25$              & $\delta\textless{}1.25^2$              & $\delta\textless{}1.25^3$              \\ \hline\hline
			\multicolumn{12}{c}{(a) Results At The Same Resolution As Training}                                                                                                                                                                                                                                                                                                                                \\ \hline
			\multicolumn{1}{l|}{Baseline}                 & \multicolumn{1}{c|}{\checkmark}                       & \multicolumn{1}{c|}{}                        & \multicolumn{1}{c|}{}                         & \multicolumn{1}{c|}{}                         & 0.107          & 0.866          & 4.680          & \multicolumn{1}{c|}{0.183}          & 0.890          & 0.963          & 0.982          \\
			\multicolumn{1}{l|}{Baseline + AS-Aug}        & \multicolumn{1}{c|}{\checkmark}                       & \multicolumn{1}{c|}{\checkmark}                   & \multicolumn{1}{c|}{}                         & \multicolumn{1}{c|}{}                         & 0.102          & 0.716          & 4.395          & \multicolumn{1}{c|}{0.178}          & 0.899          & 0.965          & 0.984          \\
			\multicolumn{1}{l|}{Baseline + D-HRNet}       & \multicolumn{1}{c|}{\checkmark}                       & \multicolumn{1}{c|}{}                        & \multicolumn{1}{c|}{\checkmark}                    & \multicolumn{1}{c|}{}                         & 0.102          & 0.746          & 4.466          & \multicolumn{1}{c|}{0.178}          & 0.897          & 0.965          & 0.983          \\
			\multicolumn{1}{l|}{RA-Depth w/o D-HRNet}     & \multicolumn{1}{c|}{\checkmark}                       & \multicolumn{1}{c|}{\checkmark}                   & \multicolumn{1}{c|}{}                         & \multicolumn{1}{c|}{\checkmark}                    & 0.097          & 0.651          & 4.246          & \multicolumn{1}{c|}{0.172}          & 0.902          & 0.967          & 0.984          \\
			\multicolumn{1}{l|}{RA-Depth w/o CS-Loss}     & \multicolumn{1}{c|}{\checkmark}                       & \multicolumn{1}{c|}{\checkmark}                   & \multicolumn{1}{c|}{\checkmark}                    & \multicolumn{1}{c|}{}                         & 0.101          & 0.732          & 4.382          & \multicolumn{1}{c|}{0.176}          & 0.900          & 0.967          & 0.984          \\
			\multicolumn{1}{l|}{\textbf{RA-Depth}} & \multicolumn{1}{c|}{\checkmark}                       & \multicolumn{1}{c|}{\checkmark}                   & \multicolumn{1}{c|}{\checkmark}                    & \multicolumn{1}{c|}{\checkmark}                    & \textbf{0.096} & \textbf{0.632} & \textbf{4.216} & \multicolumn{1}{c|}{\textbf{0.171}} & \textbf{0.903} & \textbf{0.968} & \textbf{0.985} \\ \hline
			\multicolumn{1}{l|}{Baseline w/o pt}          & \multicolumn{1}{c|}{}                            & \multicolumn{1}{c|}{}                        & \multicolumn{1}{c|}{}                         & \multicolumn{1}{c|}{}                         & 0.122          & 0.957          & 4.955          & \multicolumn{1}{c|}{0.199}          & 0.860          & 0.954          & 0.979          \\
			\multicolumn{1}{l|}{\textbf{RA-Depth} w/o pt} & \multicolumn{1}{c|}{}                            & \multicolumn{1}{c|}{\checkmark}                   & \multicolumn{1}{c|}{\checkmark}                    & \multicolumn{1}{c|}{\checkmark}                    & \textbf{0.115} & \textbf{0.806} & \textbf{4.633} & \multicolumn{1}{c|}{\textbf{0.189}} & \textbf{0.873} & \textbf{0.959} & \textbf{0.982} \\
			\hline
			\multicolumn{12}{c}{{(b) Average Results Of Resolution Adaptation At Four Different Resolutions}}\\
			\hline
			\multicolumn{1}{l|}{Baseline}                 & \multicolumn{1}{c|}{\checkmark}                       & \multicolumn{1}{c|}{}                        & \multicolumn{1}{c|}{}                         & \multicolumn{1}{c|}{}                         & 0.152          & 1.239          & 5.745          & \multicolumn{1}{c|}{0.225}          & 0.799          & 0.945          & 0.978          \\
			\multicolumn{1}{l|}{Baseline + AS-Aug}        & \multicolumn{1}{c|}{\checkmark}                       & \multicolumn{1}{c|}{\checkmark}                   & \multicolumn{1}{c|}{}                         & \multicolumn{1}{c|}{}                         & 0.107          & 0.730          & 4.470          & \multicolumn{1}{c|}{0.182}          & 0.890          & 0.964          & 0.983          \\
			\multicolumn{1}{l|}{Baseline + D-HRNet}       & \multicolumn{1}{c|}{\checkmark}                       & \multicolumn{1}{c|}{}                        & \multicolumn{1}{c|}{\checkmark}                    & \multicolumn{1}{c|}{}                         & 0.145          & 0.994          & 5.332          & \multicolumn{1}{c|}{0.220}          & 0.801          & 0.950          & 0.980          \\
			\multicolumn{1}{l|}{RA-Depth w/o D-HRNet}     & \multicolumn{1}{c|}{\checkmark}                       & \multicolumn{1}{c|}{\checkmark}                   & \multicolumn{1}{c|}{}                         & \multicolumn{1}{c|}{\checkmark}                    & 0.102          & 0.664          & 4.375          & \multicolumn{1}{c|}{\textbf{0.177}}          & 0.893          & \textbf{0.966}          & \textbf{0.985}          \\
			\multicolumn{1}{l|}{RA-Depth w/o CS-Loss}     & \multicolumn{1}{c|}{\checkmark}                       & \multicolumn{1}{c|}{\checkmark}                   & \multicolumn{1}{c|}{\checkmark}                    & \multicolumn{1}{c|}{}                         & 0.107          & 0.724          & 4.467          & \multicolumn{1}{c|}{0.182}          & 0.891          & 0.965          & 0.984          \\
			\multicolumn{1}{l|}{\textbf{RA-Depth}} & \multicolumn{1}{c|}{\checkmark}                       & \multicolumn{1}{c|}{\checkmark}                   & \multicolumn{1}{c|}{\checkmark}                    & \multicolumn{1}{c|}{\checkmark}                    & \textbf{0.101}          & \textbf{0.651}          & \textbf{4.345}          & \multicolumn{1}{c|}{\textbf{0.177}}          & \textbf{0.894}          & \textbf{0.966}          & \textbf{0.985}          \\ \hline
			\multicolumn{1}{l|}{Baseline w/o pt}          & \multicolumn{1}{c|}{}                            & \multicolumn{1}{c|}{}                        & \multicolumn{1}{c|}{}                         & \multicolumn{1}{c|}{}                         & 0.209          & 2.179          & 7.285          & \multicolumn{1}{c|}{0.290}          & 0.678          & 0.890          & 0.955          \\
			\multicolumn{1}{l|}{\textbf{RA-Depth} w/o pt} & \multicolumn{1}{c|}{}                            & \multicolumn{1}{c|}{\checkmark}                   & \multicolumn{1}{c|}{\checkmark}                    & \multicolumn{1}{c|}{\checkmark}                    & \textbf{0.124} & \textbf{0.856} & \textbf{4.844} & \multicolumn{1}{c|}{\textbf{0.198}} & \textbf{0.856} & \textbf{0.955} & \textbf{0.981} \\ 
			\bottomrule[1.5pt]
		\end{tabular}}
\end{table}

\subsection{Ablation Study}
%As show in Table~\ref{tab4}, we perform sufficient ablation experiments for each component of our proposed method.
To better comprehend how the components of our method contribute to the overall performance as well as the resolution adaptation ability, we perform sufficient ablation experiments in this section.
As shown in Table~\ref{tab_ablation}, we perform ablation studies by changing various components of our method, where all models are trained at the resolution of $640 \times 192$ on the KITTI dataset~\cite{kitti}.
%The ablation experiments in Table~\ref{tab_ablation} are mainly divided into two parts.

We conduct ablation experiments to verify  the effectiveness of each component on overall performance at the same resolution as training (640$\times$192). Moreover, we also verify the effectiveness of each component on the resolution adaptation ability at four different resolutions (416$\times$128, 512$\times$160, 832$\times$256, 1024$\times$320), where the average results are reported at four resolutions.
%One part is to test the effects of each component on overall performance at the same resolution as training (640$\times$192).
%Another part is to test the effects of each component on the resolution adaptation ability at four different resolutions (416$\times$128, 512$\times$160, 832$\times$256, 1024$\times$320), here we report the average results at the four resolutions.
For the baseline model, we use HRNet18~\cite{hrnet} as the encoder and use the same decoder as Monodepth2~\cite{monodepth2}.
The ablation results show that our proposed arbitrary-scale data augmentation (AS-Aug), Dual HRNet (D-HRNet), and cross-scale depth consistency loss (CS-Loss) can bring obvious improvements individually.
Since the CS-Loss is imposed on the AS-Aug, experiments involving the CS-Loss must include the AS-Aug operation. All combined components can yield the best performance (RA-Depth).
Besides, even without using ImageNet pre-trained weights, our model can also obtain the large gain over the baseline  model.

\begin{table}[h]
	\centering
	\caption{Additional ablation studies for AS-Aug and CS-Loss. Bian-Aug represents the data augmentation method of Bian~\etal~\cite{bian2019}. All models are trained at the resolution of $640\times192$ on the KITTI dataset~\cite{kitti}. We report the average results of resolution adaptation tested at four different resolutions (416$\times$128, 512$\times$160, 832$\times$256, 1024$\times$320).}\label{tab_ablation_aug_loss}
	\renewcommand\tabcolsep{3.0pt}
	\resizebox{0.97\textwidth}{!}{
		\begin{tabular}{l|l|cccc|ccc}
			\toprule[1.5pt]
			\multirow{2}{*}{}    & \multirow{2}{*}{}                      & \multicolumn{4}{c|}{Error Metric $\downarrow$}                                 & \multicolumn{3}{c}{Accuracy Metric $\uparrow$}                   \\ \cline{3-9} 
			&                                        & AbsRel              & SqRel              & RMSE              & RMSElog              & $\delta\textless{}1.25$              & $\delta\textless{}1.25^2$              & $\delta\textless{}1.25^3$              \\ \hline\hline
			\multirow{3}{*}{(a)} & Baseline                               & 0.152          & 1.239          & 5.745          & 0.225          & 0.799          & 0.945          & 0.978          \\
			& Baseline + Bian-Aug (1 image)          & 0.128          & 0.847          & 4.770          & 0.200          & 0.850          & 0.960          & 0.981          \\
			& Baseline + AS-Aug (1 image)            & \textbf{0.108} & \textbf{0.742} & \textbf{4.491} & \textbf{0.184} & \textbf{0.885} & \textbf{0.961} & \textbf{0.982} \\ \hline
			\multirow{3}{*}{(b)} & Baseline + AS-Aug (1 image)            & 0.108          & 0.742          & 4.491          & 0.184          & 0.885          & 0.961          & 0.982          \\
			& Baseline + AS-Aug (2 images)           & \textbf{0.107}          & 0.734          & 4.478          & \textbf{0.182}          & 0.888          & 0.963          & \textbf{0.983}          \\
			& Baseline + AS-Aug (3 images)           & \textbf{0.107} & \textbf{0.730} & \textbf{4.470} & \textbf{0.182} & \textbf{0.890} & \textbf{0.964} & \textbf{0.983} \\ \hline
			\multirow{2}{*}{(c)} & Baseline + AS-Aug (2 images) + CS-Loss & 0.103          & 0.670          & 4.383          & 0.178          & 0.891          & 0.965          & 0.984          \\
			& Baseline + AS-Aug (3 images) + CS-Loss & \textbf{0.102} & \textbf{0.664} & \textbf{4.375} & \textbf{0.177} & \textbf{0.893} & \textbf{0.966} & \textbf{0.985} \\ 
			\bottomrule[1.5pt]
		\end{tabular}}
\end{table}

\subsubsection{Effects of Arbitrary-Scale Data Augmentation.}
As shown in Table~\ref{tab_ablation}(a), our proposed arbitrary-scale data augmentation (AS-Aug) significantly improves performance when tested at the same resolution as training.
Moreover, results in Table~\ref{tab_ablation}(b) demonstrate that AS-Aug brings large gains over the baseline in terms of resolution adaptation performance.
In particular, when combining AS-Aug to the baseline model, we achieve the error reduction rates of 29.6\%, 41.4\%, and 22.2\% in terms of AbsRel, SqRel, and RMSE, respectively, and the accuracy gain of 11.4\% in terms of $\delta\textless{}1.25$.
As described in Algorithm~\ref{alg1}, given an original image, AS-Aug generates three training images with three different scales to let the model implicitly learn the scale invariance of the scene depth. When generating only one training image from an original image, we compare our AS-Aug component with the data augmentation (dubbed Bian-Aug) proposed by Bian~\etal~\cite{bian2019} as shown in Table~\ref{tab_ablation_aug_loss}(a). Although Bian-Aug also uses randomly resizing and cropping to change the scale of images, the differences between us are two-fold. On the one hand, Bian-Aug first downsizes the original image and then upsamples the downsized image to high resolution for cropping. However, our AS-Aug directly crops on the original image so that the detailed information of the images can be remained. On the other hand, \cite{bian2019} only inputs one-scale image to the network, while our method inputs three-scale images to the network so that three kinds of camera intrinsics can be simulated. By imposing the cross-scale depth consistency loss among them, we can learn the scale invariance of different image resolutions. Therefore, the proposed AS-Aug significantly outperforms the Bian-Aug in terms of resolution adaptation of depth estimation. Table~\ref{tab_ablation_aug_loss}(b) reports how the number of images generated by AS-Aug affects the resolution adaptation performance.
Since the performance is better when generating multiple images and the proposed cross-scale depth consistency loss needs to be implemented based on multiple images, we generate three training images for each input.

\begin{table}[h]
	\centering
	\caption{Parameters and GFLOPs of the depth estimation network.}\label{tab6}
	\renewcommand\tabcolsep{3.0pt}
	\resizebox{0.97\textwidth}{!}{
		\begin{tabular}{l|c|c|c|cccc|ccc}
			\toprule[1.5pt]
			\multicolumn{1}{c|}{\multirow{2}{*}{Method}} & \multirow{2}{*}{Encoder} & \multirow{2}{*}{\#Params} & \multirow{2}{*}{GFLOPs} & \multicolumn{4}{c|}{Error Metric $\downarrow$}                                 & \multicolumn{3}{c}{Accuracy Metric $\uparrow$}                   \\ \cline{5-11} 
			&                          &                             &                         & AbsRel              & SqRel              & RMSE              & RMSElog              & $\delta\textless{}1.25$              & $\delta\textless{}1.25^2$              & $\delta\textless{}1.25^3$              \\ \hline\hline
			Monodepth2~\cite{monodepth2}              & ResNet50                 & 34.57M                      & 16.63                   & 0.110          & 0.831          & 4.642          & 0.187          & 0.883          & 0.962          & 0.982          \\
			HR-Depth~\cite{HRDepth}                & ResNet18                 & 14.61M                      & 11.27                   & 0.109          & 0.792          & 4.632          & 0.185          & 0.884          & 0.962          & {\ul 0.983}    \\
			DIFF-Net~\cite{diffnet}                & HRNet18                  & {\ul 10.87M}                & 15.77                   & {\ul 0.102}    & {\ul 0.764}    & {\ul 4.483}    & {\ul 0.180}    & {\ul 0.896}    & {\ul 0.965}    & {\ul 0.983}    \\
			Baseline                & HRNet18                  & 11.31M                      & {\ul 10.99}             & 0.107          & 0.866          & 4.680          & 0.183          & 0.890          & 0.963          & 0.982          \\
			\textbf{RA-Depth}       & HRNet18                  & \textbf{9.98M}              & \textbf{10.78}          & \textbf{0.096} & \textbf{0.632} & \textbf{4.216} & \textbf{0.171} & \textbf{0.903} & \textbf{0.968} & \textbf{0.985} \\ 
			\bottomrule[1.5pt]
	\end{tabular}}
\end{table}

\subsubsection{Effects of Cross-Scale Depth Consistency Loss.}
As shown in Table~\ref{tab_ablation}, whether testing at the same resolution as training or four different resolutions, the addition of cross-scale depth consistency loss (CS-Loss) brings significant improvements.
Moreover, as shown in Table~\ref{tab_ablation_aug_loss}(c), for different settings of AS-Aug, combining CS-Loss can obviously improve the resolution adaptation performance of the model.
These experiments demonstrate that the explicit constraint of CS-Loss can further help the model learn the scale invariance of scene depth, so as to enhance the resolution adaptation ability of the model.
This verifies that our depth estimation model can explicitly learn the scale invariance of scene depth with our proposed cross-scale depth consistency loss.
%our original intention, \textit {i.e.}, learning depths from images of the same scene at different scales as well as explicitly learning the scale invariance of scene depth can help the model achieve resolution adaptation at inference.

\subsubsection{Effects of Dual HRNet.}

Table~\ref{tab6} reports the parameter complexity (\#Params) and computation complexity (GFLOPs) of our model RA-Depth.
Compared with existing state-of-the-art models as well as the baseline model, our model is the most efficient in terms of parameter complexity and computation complexity. That is, thanks to the efficient design of depth network Dual HRNet with multi-scale feature fusion, our model can achieve the best performance with the lowest network overhead.

\section{Conclusions}
\label{sec:conclusions}
We proposed RA-Depth, a self-supervised monocular depth estimation method to efficiently predict depth maps of images when the resolutions of images are inconsistent during training and testing. Based on our proposed arbitrary-scale data augmentation method, we developed a dual high-resolution network to efficiently fuse multi-scale features for the accurate depth prediction. RA-Depth achieves state-of-the-art results on various public datasets, and exhibits the good ability of resolution adaptation. Moreover, benefiting from the constructed dual HRNet, our RA-Depth model is efficient in terms of the parameter and computation complexity.
%To the best of our knowledge, this is the first work to show that a single model trained on unlabelled monocular videos can adapt well to image resolution changes at inference.

\clearpage

\bibliographystyle{splncs04}
\bibliography{egbib}

\begin{thebibliography}{10}
\providecommand{\url}[1]{\texttt{#1}}
\providecommand{\urlprefix}{URL }
\providecommand{\doi}[1]{https://doi.org/#1}

\bibitem{bhat2021adabins}
Bhat, S.F., Alhashim, I., Wonka, P.: Ada{B}ins: Depth estimation using adaptive
  bins. In: CVPR (2021)

\bibitem{bian2019}
Bian, J.W., Li, Z., Wang, N., Zhan, H., Shen, C., Cheng, M.M., Reid, I.:
  Unsupervised scale-consistent depth and ego-motion learning from monocular
  video. In: NeurIPS (2019)

\bibitem{chen2019towards}
Chen, P.Y., Liu, A.H., Liu, Y.C., Wang, Y.C.F.: Towards scene understanding:
  Unsupervised monocular depth estimation with semantic-aware representation.
  In: CVPR (2019)

\bibitem{safenet}
Choi, J., Jung, D., Lee, D., Kim, C.: Safenet: Self-supervised monocular depth
  estimation with semantic-aware feature extraction. arXiv preprint
  arXiv:2010.02893  (2020)

\bibitem{eigen2014depth}
Eigen, D., Puhrsch, C., Fergus, R.: Depth map prediction from a single image
  using a multi-scale deep network. arXiv preprint arXiv:1406.2283  (2014)

\bibitem{fu2018deep}
Fu, H., Gong, M., Wang, C., Batmanghelich, K., Tao, D.: Deep ordinal regression
  network for monocular depth estimation. In: CVPR (2018)

\bibitem{garg2016unsupervised}
Garg, R., Bg, V.K., Carneiro, G., Reid, I.: Unsupervised cnn for single view
  depth estimation: Geometry to the rescue. In: ECCV (2016)

\bibitem{kitti}
Geiger, A., Lenz, P., Stiller, C., Urtasun, R.: Vision meets robotics: The
  kitti dataset. The International Journal of Robotics Research  (2013)

\bibitem{monodepth}
Godard, C., Mac~Aodha, O., Brostow, G.J.: Unsupervised monocular depth
  estimation with left-right consistency. In: CVPR (2017)

\bibitem{monodepth2}
Godard, C., Mac~Aodha, O., Firman, M., Brostow, G.J.: Digging into
  self-supervised monocular depth estimation. In: ICCV (2019)

\bibitem{packnet}
Guizilini, V., Ambrus, R., Pillai, S., Raventos, A., Gaidon, A.: 3{D} packing
  for self-supervised monocular depth estimation. In: CVPR (2020)

\bibitem{guizilini2020semantically}
Guizilini, V., Hou, R., Li, J., Ambrus, R., Gaidon, A.: Semantically-guided
  representation learning for self-supervised monocular depth. arXiv preprint
  arXiv:2002.12319  (2020)

\bibitem{resnet}
He, K., Zhang, X., Ren, S., Sun, J.: Deep residual learning for image
  recognition. In: CVPR (2016)

\bibitem{sgm2005}
Hirschmuller, H.: Accurate and efficient stereo processing by semi-global
  matching and mutual information. In: CVPR (2005)

\bibitem{sgm2007}
Hirschmuller, H.: Stereo processing by semiglobal matching and mutual
  information. IEEE Transactions on Pattern Analysis and Machine Intelligence
  (2007)

\bibitem{hu2019revisiting}
Hu, J., Ozay, M., Zhang, Y., Okatani, T.: Revisiting single image depth
  estimation: Toward higher resolution maps with accurate object boundaries.
  In: WACV (2019)

\bibitem{stn}
Jaderberg, M., Simonyan, K., Zisserman, A., Kavukcuoglu, K.: Spatial
  transformer networks. arXiv preprint arXiv:1506.02025  (2015)

\bibitem{kingma2014adam}
Kingma, D.P., Ba, J.: Adam: A method for stochastic optimization. arXiv
  preprint arXiv:1412.6980  (2014)

\bibitem{sgddepth}
Klingner, M., Term{\"o}hlen, J.A., Mikolajczyk, J., Fingscheidt, T.:
  Self-supervised monocular depth estimation: Solving the dynamic object
  problem by semantic guidance. In: ECCV (2020)

\bibitem{ladicky2014pulling}
Ladicky, L., Shi, J., Pollefeys, M.: Pulling things out of perspective. In:
  CVPR (2014)

\bibitem{laina2016deeper}
Laina, I., Rupprecht, C., Belagiannis, V., Tombari, F., Navab, N.: Deeper depth
  prediction with fully convolutional residual networks. In: 3DV (2016)

\bibitem{lee2019big}
Lee, J.H., Han, M.K., Ko, D.W., Suh, I.H.: From big to small: Multi-scale local
  planar guidance for monocular depth estimation. arXiv preprint
  arXiv:1907.10326  (2019)

\bibitem{leistner2022towards}
Leistner, T., Mackowiak, R., Ardizzone, L., K{\"o}the, U., Rother, C.: Towards
  multimodal depth estimation from light fields. In: CVPR (2022)

\bibitem{liu2015deep}
Liu, F., Shen, C., Lin, G.: Deep convolutional neural fields for depth
  estimation from a single image. In: CVPR (2015)

\bibitem{liu2015learning}
Liu, F., Shen, C., Lin, G., Reid, I.: Learning depth from single monocular
  images using deep convolutional neural fields. IEEE Transactions on Pattern
  Analysis and Machine Intelligence  (2015)

\bibitem{luo2019every}
Luo, C., Yang, Z., Wang, P., Wang, Y., Xu, W., Nevatia, R., Yuille, A.: Every
  pixel counts++: Joint learning of geometry and motion with 3{D} holistic
  understanding. IEEE Transactions on Pattern Analysis and Machine Intelligence
   (2019)

\bibitem{HRDepth}
Lyu, X., Liu, L., Wang, M., Kong, X., Liu, L., Liu, Y., Chen, X., Yuan, Y.:
  Hr-depth: High resolution self-supervised monocular depth estimation. AAAI
  (2020)

\bibitem{3dunsupervised}
Mahjourian, R., Wicke, M., Angelova, A.: Unsupervised learning of depth and
  ego-motion from monocular video using 3{D} geometric constraints. In: CVPR
  (2018)

\bibitem{parida2021beyond}
Parida, K.K., Srivastava, S., Sharma, G.: Beyond image to depth: Improving
  depth prediction using echoes. In: CVPR (2021)

\bibitem{pytorch}
Paszke, A., Gross, S., Chintala, S., Chanan, G., Yang, E., DeVito, Z., Lin, Z.,
  Desmaison, A., Antiga, L., Lerer, A.: Automatic differentiation in pytorch
  (2017)

\bibitem{patil2022p3depth}
Patil, V., Sakaridis, C., Liniger, A., Van~Gool, L.: {P3D}epth: Monocular depth
  estimation with a piecewise planarity prior. In: Proceedings of the IEEE/CVF
  Conference on Computer Vision and Pattern Recognition. pp. 1610--1621 (2022)

\bibitem{epcdepth}
Peng, R., Wang, R., Lai, Y., Tang, L., Cai, Y.: Excavating the potential
  capacity of self-supervised monocular depth estimation. In: ICCV (2021)

\bibitem{Poggi_CVPR_2020}
Poggi, M., Aleotti, F., Tosi, F., Mattoccia, S.: On the uncertainty of
  self-supervised monocular depth estimation. In: CVPR (2020)

\bibitem{poggi2018learning}
Poggi, M., Tosi, F., Mattoccia, S.: Learning monocular depth estimation with
  unsupervised trinocular assumptions. In: 3DV (2018)

\bibitem{qi2018geonet}
Qi, X., Liao, R., Liu, Z., Urtasun, R., Jia, J.: Geo{N}et: Geometric neural
  network for joint depth and surface normal estimation. In: CVPR (2018)

\bibitem{ranjan2019competitive}
Ranjan, A., Jampani, V., Balles, L., Kim, K., Sun, D., Wulff, J., Black, M.J.:
  Competitive collaboration: Joint unsupervised learning of depth, camera
  motion, optical flow and motion segmentation. In: CVPR (2019)

\bibitem{imagenet}
Russakovsky, O., Deng, J., Su, H., Krause, J., Satheesh, S., Ma, S., Huang, Z.,
  Karpathy, A., Khosla, A., Bernstein, M., et~al.: Imagenet large scale visual
  recognition challenge. International journal of computer vision  (2015)

\bibitem{make3d}
Saxena, A., Sun, M., Ng, A.Y.: Make3{D}: Learning 3{D} scene structure from a
  single still image. IEEE transactions on pattern analysis and machine
  intelligence  (2008)

\bibitem{nyuv2}
Silberman, N., Hoiem, D., Kohli, P., Fergus, R.: Indoor segmentation and
  support inference from rgbd images. In: ECCV (2012)

\bibitem{hrnet}
Wang, J., Sun, K., Cheng, T.: Deep high-resolution representation learning for
  visual recognition. IEEE Transactions on Pattern Analysis and Machine
  Intelligence  (2019)

\bibitem{ssim}
Wang, Z., Bovik, A.C., Sheikh, H.R., Simoncelli, E.P.: Image quality
  assessment: from error visibility to structural similarity. IEEE transactions
  on image processing  (2004)

\bibitem{depthhints}
Watson, J., Firman, M., Brostow, G.J., Turmukhambetov, D.: Self-supervised
  monocular depth hints. In: ICCV (2019)

\bibitem{xu2018structured}
Xu, D., Wang, W., Tang, H., Liu, H., Sebe, N., Ricci, E.: Structured attention
  guided convolutional neural fields for monocular depth estimation. In: CVPR
  (2018)

\bibitem{yin2019enforcing}
Yin, W., Liu, Y., Shen, C., Yan, Y.: Enforcing geometric constraints of virtual
  normal for depth prediction. In: ICCV (2019)

\bibitem{IndoorSfMLearner}
Yu, Z., Jin, L., Gao, S.: P$^{2}$net: Patch-match and plane-regularization for
  unsupervised indoor depth estimation. In: ECCV (2020)

\bibitem{zhan2018unsupervised}
Zhan, H., Garg, R., Weerasekera, C.S., Li, K., Agarwal, H., Reid, I.:
  Unsupervised learning of monocular depth estimation and visual odometry with
  deep feature reconstruction. In: CVPR (2018)

\bibitem{zhang2019pattern}
Zhang, Z., Cui, Z., Xu, C., Yan, Y., Sebe, N., Yang, J.: Pattern-affinitive
  propagation across depth, surface normal and semantic segmentation. In: CVPR
  (2019)

\bibitem{diffnet}
Zhou, H., Greenwood, D., Taylor, S.: Self-supervised monocular depth estimation
  with internal feature fusion. In: BMVC (2021)

\bibitem{zhou2017}
Zhou, T., Brown, M., Snavely, N., Lowe, D.G.: Unsupervised learning of depth
  and ego-motion from video. In: CVPR (2017)

\bibitem{zhu2020edge}
Zhu, S., Brazil, G., Liu, X.: The edge of depth: Explicit constraints between
  segmentation and depth. In: CVPR (2020)

\end{thebibliography}
\end{document}